\def\year{2017}\relax
\documentclass[letterpaper]{article}
\usepackage{aaai17}
\usepackage{times}
\usepackage{helvet}
\usepackage{courier}
\usepackage{xr}

\usepackage{graphicx}
\usepackage[algoruled,linesnumbered]{algorithm2e}

\usepackage{multirow}
\usepackage{amsmath}
\usepackage{amsfonts}
\usepackage{amsthm}
\usepackage{amstext}

\newtheorem{theorem}{Theorem}[section]

\allowdisplaybreaks[3]

\begin{document}
%

\title{Phase-Mapper: An AI Platform to \\Accelerate High Throughput Materials Discovery}
\author{
 Yexiang	Xue\\		
 Department of Computer Science\\
 Cornell University\\
 yexiang@cs.cornell.edu 
 \And
 Junwen Bai \\
 Zhiyuan College\\
 Shanghai Jiao Tong University, China\\
 bjw\_sjtu@sjtu.edu.cn
 \And
 Ronan Le Bras,  Brendan Rappazzo\\ 
 Department of Computer Science \\
 Cornell University\\
 \{rl454, bhr54\}@cornell.edu
 \AND
 Richard Bernstein, Johan Bjorck, Liane Longpre
\\ 
 Department of Computer Science \\
 Cornell University\\
 \{rab38, njb225, lfl42\}@cornell.edu
 \And
 Santosh K. Suram \\  Joint Center for Artificial Photosynthesis
 \\ California Institute of Technology
 \\ sksuram@caltech.edu  
 \AND
 Robert B. van Dover \\ Materials Science and Engineering
 \\ Cornell University \\ rbv2@cornell.edu
 \And
 John Gregoire \\  Joint Center for Artificial Photosynthesis
 \\ California Institute of Technology
 \\ gregoire@caltech.edu  
 \And
 Carla P. Gomes \\  Department of Computer Science 
 \\ Cornell University \\ gomes@cs.cornell.edu  
}
\maketitle
\begin{abstract}



High-Throughput materials discovery involves the rapid synthesis, measurement, and characterization of many different but structurally-related materials. A key problem in materials discovery, the phase map identification problem, involves the determination of the crystal phase diagram from the materials’ composition and structural characterization data.  We present Phase-Mapper, a novel AI platform to solve the phase map identification problem that allows humans to interact with both the data and products of AI algorithms, including the incorporation of human feedback to constrain or initialize solutions. Phase-Mapper affords incorporation of any spectral demixing algorithm, including our novel solver, AgileFD, which is based on a convolutive non-negative matrix factorization algorithm. AgileFD can incorporate constraints to capture the physics of the materials as well as human feedback. We compare three solver variants with previously proposed methods in a large-scale experiment involving 20 synthetic systems, demonstrating the efficacy of imposing physical constrains using AgileFD. Phase-Mapper has also been used by materials scientists to solve a wide variety of phase diagrams, including the previously unsolved Nb-Mn-V oxide system, which is provided here as an illustrative example.
\end{abstract}



\section{Introduction}

\begin{figure}[tb]
  \centering
  \includegraphics[width=0.95\linewidth]{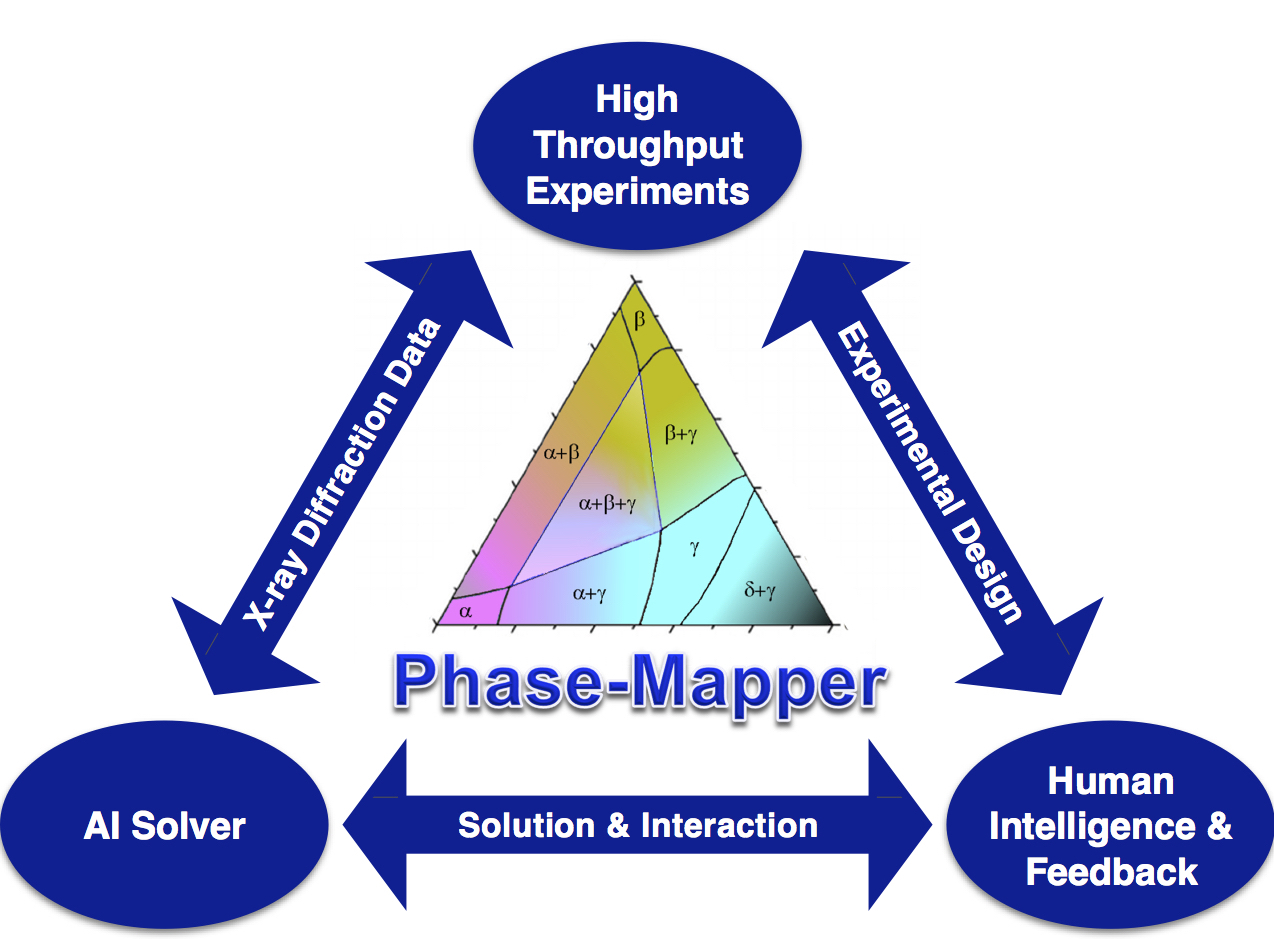}
  \caption{The Phase-Mapper Platform integrates experimentation,AI solvers and human feedback into a platform for High Throughput Materials Discovery for discovering new materials.}
  \vspace{-10pt}
  \label{fig:triangle}
\end{figure}

The wonders of modern technology can largely be attributed to advances in materials science that enable innovations from semiconductors to renewable energy. High throughput materials discovery comprises a suite of emerging methodologies to rapidly identify new materials, especially, un-discovered materials that are critical for next-generation technologies~\cite{Green2013}. Specifically, this method generates 10$^2$-10$^3$ unique materials and forms them into a ``library'' and then rapidly screens for the properties of interest. 

To analyze the vast amount of data that are generated in high throughput experiments, automatic analysis becomes imperative. The traditional workflow in materials science relies on iterative manual analysis and heuristics, resulting in months or years of analysis for a single physical system. This quickly becomes a bottleneck as humans are unable to keep up with the rate at which data is generated from high throughput experiments. The need for automatic and scalable tools in materials science provides computer scientists unique opportunities to apply cutting edge techniques in Artificial Intelligence and data science to accelerate the materials discovery process.

%
In this paper we address the \textit{phase mapping problem}, a central issue in high-throughput materials discovery, and provide an efficient method of solving it which is currently a critically missing component in the high throughput materials toolbox.  A material’s phase is a fundamental property of a material composition (a mixture of elements) that describes the arrangement of the constituent atoms. X-ray diffraction (XRD) is a ubiquitous technique to characterize phases, as it produces a signal containing a series of peaks that serve as a ``fingerprint'' of the underlying atomic arrangement (or crystal structure). Using traditional methods, materials scientists can obtain and interpret 1-10 XRD measurements per day, and with the recent development of automated, synchrotron-based XRD experiments, the measurement throughput has been accelerated to 10$^3$-10$^5$ measurements per day~\cite{Gregoire2009,Gregoire2014}. The creation of a phase mapping algorithm that generates phase diagrams from these data remains an unsolved problem in materials science despite a series of advancements over the past decade~\cite{Hattrick-simpers2016}.  The most pertinent need is to generate a physically-meaningful phase diagram for the 10$^2$-10$^3$ materials in a given library, which relies on the spectral demixing of the 10$^2$-10$^3$ XRD patterns into a small set of basis patterns (typically less than 10).

To address this substantial challenge, we developed {\bf Phase-Mapper}, a comprehensive AI platform that tightly integrates XRD experimentation, AI problem solving, and human intelligence for the phase mapping problem (See Figure~\ref{fig:triangle}). In this platform, \textit{{\bf within minutes}}, AI solvers give approximate results for the phase mapping problem, which then are examined and further refined by materials scientists interactively and in real time. In addition, the results of Phase-Mapper can be used to further inform future experimental designs. The demixing algorithm is a cornerstone of the Phase-Mapper platform, and we have developed a {\bf novel solver called AgileFD}, based on Convolutive Non-negative Matrix Factorization (cNMF), a method that has been applied to blind source separation of audio signals and speech recognition~\cite{smaragdis2004non,morup2006sparse}. 

AgileFD {\bf features a lightweight iterative updates of candidate solutions}, and we have developed a suite of adaptations that enable functionalities beyond cNMF. The extensions for AgileFD described here include incorporation of constraints to encode both {\bf human input}, that capitalizes on a researcher’s knowledge of a particular dataset, and \textit{a priori} knowledge of the problem related to the {\bf underlying physics} of phase diagrams.This, as demonstrated below, can be critical in obtaining physically meaningful solutions. In developing the Phase-Mapper platform, careful attention has been given to deliver a rich suite of capabilities while maintaining solver convergence times within minutes, which enables researchers to interact with the solver to refine the solution.
%

We compare three variants of the AgileFD with previously proposed solvers on {\bf an experiment involving 20 synthetically generated systems}. Our results show that AgileFD outperforms previous solver in terms of reconstruction error and model correctness, owing to the fact that the model represents peak shifting in an efficient way. In addition, light-weight update rules allow AgileFD to converge more quickly than previous solvers and with its extensions the obtained solutions tend to be more physically meaningful. 

Phase-Mapper has been deployed in the Joint Center for Artificial Photosynthesis (JCAP) supported by the Department of Energy for materials scientists to solve a wide variety of phase diagrams, including the {\bf previously unsolved Nb-Mn-V oxide system}, which is provided here as a case study and an illustrative example of the importance of encoding physical constraints to obtain physically-meaningful phase diagram solutions.

\section{Phase-Mapper: AI for Materials Discovery}


\begin{figure}[tb]
  \centering
  \includegraphics[width=0.95\linewidth]{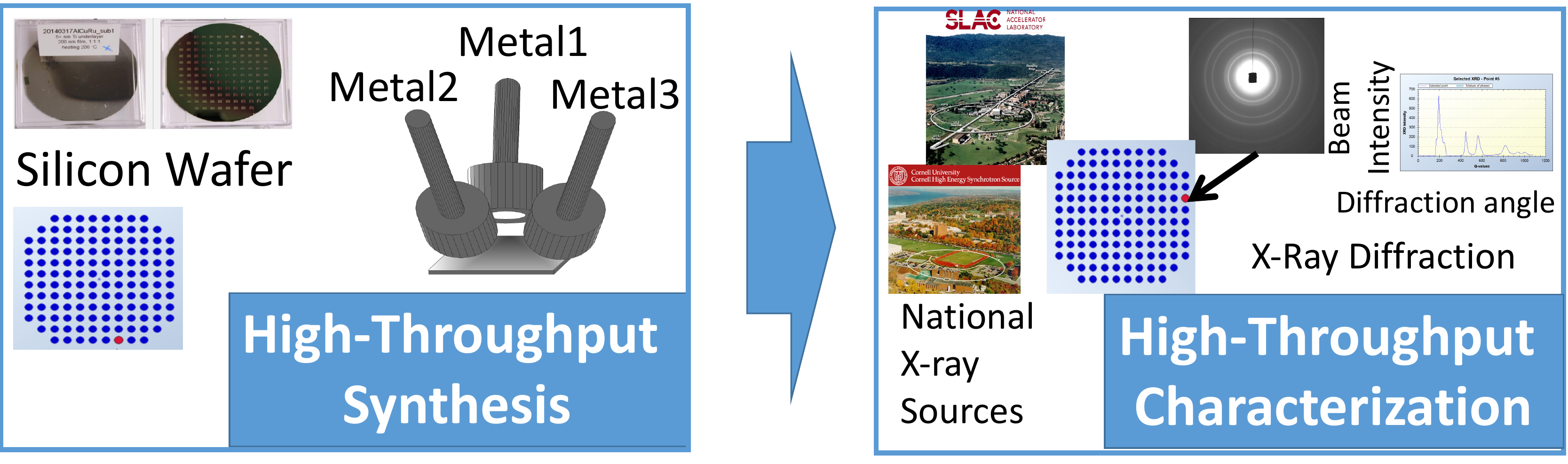}
  \caption{High throughput synthesis and characterization of materials.}
  \vskip -15pt
  \label{fig:phy_exp}
\end{figure}


High throughput materials discovery is an experimentation pipeline for rapidly synthesizing and identifying new materials. 
In this pipeline, a handful of elements are deposited together on a two-dimensional substrate, so that different locations on the substrate receive varying proportions of the elements.
This variation in elemental composition across the substrate gives rise to the forming of a discrete set of materials.
Each material is present on particular segments of the substrate, due to the continuous change of the proportion of the elements across the substrate.

After deposition, sample locations on the substrate are probed with high energy X-rays, and because each material has a characteristic XRD pattern it becomes possible to characterize the discrete set of materials present on the substrate. Figure~\ref{fig:phy_exp} provides a graphical illustration of the synthesis and characterization step in the high throughput pipeline.

Unlike in traditional powder diffraction methods for materials discovery, the XRD patterns obtained in the high throughput pipeline can be composed of the characteristic XRD patterns of several materials. Therefore, the \textit{phase-mapping problem}, which is to identify the characteristic XRD patterns for the materials (or basis patterns) that de-mix the signal, lies at the heart of the analysis of high throughput data. Mathematically, the measured XRD pattern in the $j$-th sample point can be characterized by a one dimensional signal $A_j(q)$. The ``scattering vector magnitude'' $q$ is a monotonic transformation of the diffraction angle, and is directly related to the spacing of atoms in a crystal. The phase-mapping problem is to find a small number of phases $W_1(q), \ldots, W_K(q)$, such that the XRD patterns at each sample point can be explained by a linear combination of phases:
\begin{align}
A_j(q) \approx \sum_{i=1}^K h_{ij} W_i(\lambda_{ij} q ). \qquad \forall j \label{eq:phasemap}
\end{align}
In the above definition, we write $W_i(\lambda_{ij} q )$ to allow for the phases to scale slightly according to parameter $\lambda_{ij}$ at each sample point.  This is the result of a commonly observed form of alloying, a process that can typically be approximated by a multiplicative scaling of the XRD pattern of a specific phase in the $q$ domain.
We also call this process peak shifting, because the effect appears to make XRD patterns in the data shift to the left or to the right.
In addition to the complications introduced by peak shifting, there are a number of other constraints on the solution of the phase-mapping problem, arising from the fact that the solution must describe a system constrained by the laws of physics. The most prominent is the so called Gibbs phase rule, which states that no more than three coefficients among $h_{ij}$ for fixed $j$ may be nonzero (in a ternary system). Additionally, how the $h_{ij}$ may vary spatially on the substrate, as well as the shapes that $W_i$ may take, are constrained by physical laws.

Fundamentally novel techniques are required to solve the phase mapping problem quickly and accurately. A number of automatic techniques have been developed in recent years, which can be broadly grouped into clustering, constraint reasoning, and factor decomposition approaches. Proposed clustering methods such as hierarchical clustering (HCA) \cite{long07}, dynamic time warping kernel clustering~\cite{LeBras11} and mean shift theory~\cite{kusne14} produce maps of phase regions, but fail to resolve mixtures or identify basis patterns, and do not necessarily produce results consistent with physics. Constraint reasoning approaches include satisfiability modulo theory (SMT) methods~\cite{ermon2012smt}, which can provide physically meaningful results, but depend heavily on effective pre-processing such as peak identification, and are computationally intensive. Approaches based on non-negative matrix factorization (NMF)~\cite{long09} are computationally efficient, but generally perform poorly when peak-shifting phenomena are present. Another factor decomposition approach called CombiFD \cite{combiFD15} uses combinatorial constraints to simultaneously enforce physical rules and accommodate peak shifting, but requires solving a combinatorial problem in each descent step, and is therefore computationally expensive.

We propose the AI platform Phase-Mapper, which integrates three key components to solve the problem: (i) cutting edge AI solvers; (ii) human intelligence and feedback; and (iii) high-throughput physical experiments 
(See Figure~\ref{fig:triangle}). 
\begin{itemize}
\item Our platform is supported by cutting edge AI solvers, including
  Non-negative Matrix Factorization \cite{lee2001algorithms}, and CombiFD. We
  also highlight a new solver called AgileFD as a key component of the
  platform. Motivated by Convolutive NMF, AgileFD features a set of
  light-weight updating rules and therefore a very fast gradient
  descent process.
AgileFD is also flexible, allowing for additional physical constraints or incorporation of human feedback through rounding and refinement.
\item Our platform also provides tools for data exploration, visualization, and configuration that allows human experts as well as laypeople to analyze and improve solutions.
\item The solutions obtained by the interaction between solvers and human users could also shed light on the development of new physical experiments, for example by specifying regions of composition space to sample at higher resolution (active learning). Therefore, the three components of Phase-Mapper form an integrated process (see Figure~\ref{fig:triangle}).
\end{itemize}

 \section{AgileFD as a Novel Solver}


AgileFD is a key component in the Phase-Mapper platform. Compared with previously proposed methods for solving the phase-mapping problem, AgileFD features quick iterative updates of  candidate solutions, which makes it possible for human experts to interact with the algorithm in real time. The key behind this speed lies in the efficient problem representation. Let the XRD patterns for all samples be represented by a matrix $A$ where each column corresponds to one sample point and each row corresponds to $A_j(q)$ for a particular value of $q$. Under the assumptions of no noise and no shifting, i.e. for all $i,j$, $\lambda_{ij} = 1$, describing $A$ as a linear combination of a few basis patterns $W_i(q)$ is equivalent to factorizing $A$ as a product of two low rank matrices $W$ and $H$. We enforce nonnegativity for $W$ and $H$, which is required for the solutions to be physically meaningful.  

\begin{displaymath}
\begin{aligned}
A \approx W \cdot H.
\end{aligned}
\end{displaymath}

In this formulation, the columns of $W$ form a set of basis patterns $W_i(q)$, while the columns of $H$ corresponds to the values $h_{ij}$ in equation \ref{eq:phasemap}. Previous approaches to solve the phase-mapping problem based on NMF have been unsuccessful in handling peak shifting, i.e. $\lambda_{ij} \neq 1$. The first contribution of AgileFD is to circumvent the shifting problem by a log space resampling. Under the variable substitution $q \rightarrow \log q$ our signal becomes $W_i(\log q)$. More importantly, the shifted phase $W_i(\log \lambda q)$ becomes $W_i(\log \lambda + \log q)$, which transforms the multiplicative shift in the $q$ domain into a constant additive offset. Once a solution is found, the transformation is reversed, and the solution is interpolated at the original $q$ values. This allows the problem to be formulated in terms of convolutive nonnegative matrix factorization. After this variable substitution, we discretize the values of allowed $\lambda$ and interpolate the signals at the corresponding geometric series $q$ values. The problem can then be written:

\begin{align}
A \approx R = \sum\limits_m {W}^{\downarrow^m} \cdot H^m.\label{eq:conv_nmf} 
\end{align}
Here, the columns of $W$ still represent basis
patterns. $W^{\downarrow m}$ is the result of shifting the rows of the
$W$ matrix down $m$ rows, and padding the shifted $m$ rows with 0, 
representing the basis patterns with a constant offset in the $log q$ domain, which is equivalent to the original multiplicative shift in the $q$ domain. The columns of $H^m$ act as the activation of basis patterns for the basis patterns shifted down $m$ units. Note that when $M=1$, this formulation is equivalent to NMF aside from the $\log$ transformation.

AgileFD is a family of algorithms, which can be adapted to use different loss functions, regularization, and certain imposed constraints.
Equation (\ref{eq:conv_nmf}) is adapted from Convolutive NMF (cNMF), which was first proposed to analyze audio signals~\cite{smaragdis2004non}. The phase-mapping problem differs from previous applications of cNMF for blind source separation as the $log q$ domain is substituted for the time domain, and each source (phase) is expected to appear at most once per sample with a relatively small offset. As in cNMF, we use a gradient descent approach to fit $W$ and $H$. The gradient updates can typically be written multiplicatively, and are applied iteratively until convergence. For example, when the generalized Kullback-Leibler divergence is chosen as the loss function, the following update rules can be applied: 
\begin{align}
 H^{m} &=H^{m} \circ \dfrac{({W}^{\downarrow m})^T ({A}/{R})}{{({W^{\downarrow m}})}^T \mathbf{1}},\label{eq:kl_update_H}\\
 W &= W \circ \dfrac{\sum\limits_{m}\left(\dfrac{{A}^{\uparrow m}}{{R}^{\uparrow m}}\right)\left({H^{m}}\right)^T}{\sum\limits_{m}\mathbf{1}\left({H^{m}}\right)^T}.\label{eq:kl_update_W}
\end{align}
Here, ${A}^{\uparrow m}$ denotes $A$ with all rows shifted upwards $m$ steps and padded with zeros, and $\mathbf{1}$ is the ones matrix. All divisions the in above expressions are taken elementwise, and $\circ$ denotes the Hadamard product.




\subsubsection{Lightweight Update Rules}
AgileFD's gradient descent method with simple linear update rules results in very fast convergence, typically reaching local minima within minutes or faster. This runtime performance is orders of magnitude faster than CombiFD, which uses a similar problem formulation but with combinatorial constraints enforced. This increased efficiency enables high throughput analysis and also makes it possible for a human to interact with the system almost in real time.

\subsubsection{Further Extensions of AgileFD for Materials Discovery}
Since the ultimate aim of the phase-mapping problem is to find a physically meaningful decomposition of the signal, for which the loss function is just a proxy, we must allow for experts to modify and inspect any candidate solution. It is not feasible to encode all physical constraints or the knowledge of a materials scientist within the solver a priori. Therefore, in the next few sections, we provide a number of novel modifications to the basic AgileFD algorithm, in order to impose prior knowledge or additional constraints derived from user interpretation of a proposed solution.

\subsubsection{AgileFD with Frozen Values}

In the Phase-Mapper platform, the user is provided with the opportunity to freeze individual values in the $W$ and $H$ matrices. For example, a user might specify a known pattern or part of a previous solution as a basis pattern a priori, freezing the corresponding row or part thereof in $W$ to a known value. Or the user might specify that a certain set of samples contain only a single material phase and set the non-corresponding $H$ vales to zero. The result is an interactive, iterative matrix factorization. We adapt the update rules to the this new scenario. 
\begin{theorem}
  When the generalized KL Divergence is chosen as the loss function,
  the objective is always non-negative and non-increasing when
  applying the update rules in Equation~\ref{eq:kl_update_H} and
  Equation~\ref{eq:kl_update_W} while holding one or more basis
  patterns in $W$ or phase proportions in $H$ fixed.
 \label{thm:update_rule_freeze}
\end{theorem}
The proof of Theorem~\ref{thm:update_rule_freeze} can be found in \cite{anon2016}. It relies on the fact that the auxilliary function used to prove that the normal update rules are non-increasing has certain linearity properties.

\subsubsection{Custom Initialization}

The local optimum to which AgileFD converges is not guaranteed to be the correct optimum. In this case, initializing basis patterns or coefficients to values close to the expected solution, rather than random values, the user can direct the search to the correct solution space. We allow the user to specify basis patterns that can be taken from previous solutions, data samples, or provided manually, to use as an initial value. Similarly, initial values for the activation matrix can be specified.

\subsubsection{Sparsity Regularization}

Sparse solutions are usually more easily interpreted, and in materials science they are more likely to be consistent with the underlying physics. The AgileFD system provides the option to introduce a sparsity term in $H$ which can vary by index according to an human experts preferences. Using L1-regularization for $H$ and sparsity weight matrices $\gamma_m$, the sparse generalized KL-divergence objective function becomes:
\begin{align*}
\sum\limits_{i,j}\left(A_{i,j}\log \dfrac{A_{i,j}}{R_{i,j}}-R_{i,j}+A_{i,j}\right) + \sum_m  \| \gamma_m \circ H^m\|_1.
\end{align*}
The corresponding update rule for $H$ becomes:
\begin{align*}
 H^{m} &=H^{m} \circ \dfrac{({W}^{\downarrow m})^T ({A}/{R})}{{({W^{\downarrow m}})}^T \mathbf{1} + \gamma_m}. 
\end{align*}
In order to avoid the degenerate solution where $H \rightarrow 0$, each basis pattern of $W$ is L2-normalized at the beginning of each update iteration.

\subsubsection{Sparsity Constraints}
In general, correct phase map solutions should follow the Gibbs phase rule, which specifies that under certain experimental conditions, the number of observed phases at a given chemical composition is less than or equal to the number of chemical elements $N_{el}$ present:
\begin{align}
\sum_{i}I_{ij} \le N_{el}. \label{eq:gibbs_phase_rule} 
\end{align}
Here, $I_{ij}$ is an indicator of whether phase $i$ is present at sample location $j$. 
Materials scientists might also know a priori, or infer from previous proposed solutions, that certain regions contain fewer phases than the usual limit. 

Such combinatorial constraints cannot be encoded directly in the update rules of AgileFD. One way to enforce these constraints, which has been used in previous methods such as CombiFD \cite{combiFD15}, is to directly encode it as hard combinatorial constraints. However, this results in a very slow update process, as we have to solve a Mixed Integer Programming (MIP) problem in each iteration. As a novel routine, we apply the Gibbs Phase rule by first solving the relaxed problem, then we choose the best values to set to zero in $H$ to satisfy  Equation~\ref{eq:gibbs_phase_rule}, and then refine the solution by applying the update rules until convergence. Because the update rules are multiplicative, the zeroed values will remain zero.

Choosing the values to zero in $H$ is independent for each sample point $j$. This can be solved greedily if a faster solution is desired, or using a MIP formulation, and/or successive rounds of constraints and refinement, if a more precise solution is desired with a somewhat longer wait time.  
This extension is particularly useful when the unconstrained algorithm recovers a solution that is nearly correct except for relatively small violations of phase limits.

\section{Human Integrated Platform}

\begin{figure}[bt]
  \centering
  \includegraphics[width=0.9\linewidth]{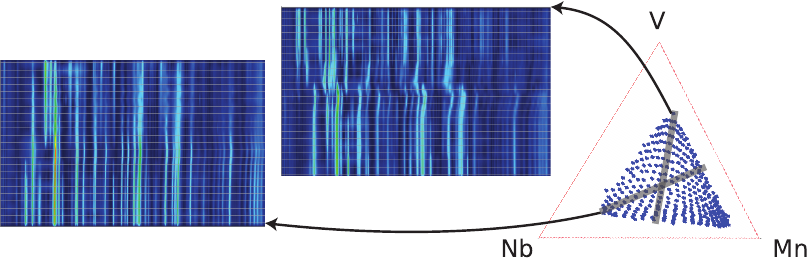}
  \vskip -10pt
  \caption{Two heatmaps of XRD patterns generated by taking slices in the visualizer. }
  \label{fig:heatMap}
\end{figure} 

We provide an integrated workflow with the Phase-Mapper platform, which includes visualizing and analyzing an instance file, setting the solver framework, analyzing the solution and using that analysis to update the solver framework. The design objectives were simple: create a practical application that seamlessly connects a visualization system with a powerful solver, that could be utilized in a large scale manner. The main features of Phase-Mapper are the visualization tools and the solver interface. 

\subsubsection{Visualizer} Phase-Mapper provides a way to visualize both the input data as well as the solution that is generated. When an instance file of a materials system is uploaded to the system, the visualizer will generate \textit{a composition map}, which illustrates the varying compositions, of elements, for all sample points. The user can freely inspect  the XRD patterns of each sample point, as well as the heatmap of XRD patterns for a slice of sample points.
A slice heatmap example is shown in Figure \ref{fig:heatMap}, where two selected
\textit{slices} of sample points are shown in grey. The heatmap on
the left represents the XRD patterns at the sample points in the slice. 
%
The heatmap also allows for multiple views of the data, emphasizing
different parts of the XRD signal, as well as a way to freeze the
current heatmap.


When the solution files are loaded to the application, either uploaded by the user or generated by the built in solver, several new plots are generated. The first of which, plots the basis patterns that were found as solutions. The second is another composition map, displaying how much each basis pattern contributed to each sample point's signal. The third plot shows the actual XRD signal as well as the reconstructed signal for a user specified sample point. Additionally the user can toggle that plot so it shows each basis pattern's contribution to the reconstructed signal. 


\subsubsection{Connection to Solver} The solving feature of Phase Mapper enables users to interact with the AI solver behind the scenes. 
The user can specify many solver parameters such as how much to enforce sparsity, how many phases the solution should have, and how much shift between basis patterns it should allow. Perhaps the most useful parameter that can be set is the initilization or freezing of a particular sample point's XRD pattern as a phase in the actual solution. The user can select a single data point on the composition map and use it’s XRD pattern as an initialization basis for the solver, or freeze that pattern as one of the basis patterns.
The former is ideal for when the user believes a data point's XRD pattern is similar to a basis pattern solution. The latter is best for when the user believes a data point's XRD pattern is a basis pattern solution. Both input parameters aim to help the solver more efficiently and accurately find a solution, either by starting the solver off closer to a solution, or distorting the solution space so the solver finds a more accurate solution.

%


\subsubsection{Infrastructure } It was pertinent that this tool be highly accessible, and for this reason a web application was chosen as the medium to deploy this project. This application runs on any modern web browser and for that reason is completely cross-platform. On the front end, HTML and CSS were used to generate the aesthetics and structure for how the different components were presented to the user. JavaScript handled all of the client side visualization and functionality. Jquery and Ajax were used to handle the client-server communication, and all server side functionality was given by PHP. The solution parameters were passed to the solver using PHP and solver itself was written in c++. Little maintence of the platform is needed as the methods in materials science as not expected to change, and as the intended audience is well versed in the phase mapping problem there is scarce need for tutorials.

\section{Experiments}


\subsection{Large Scale Experiments}

\begin{figure}[b]
  \centering
  \includegraphics[width=0.7\linewidth]{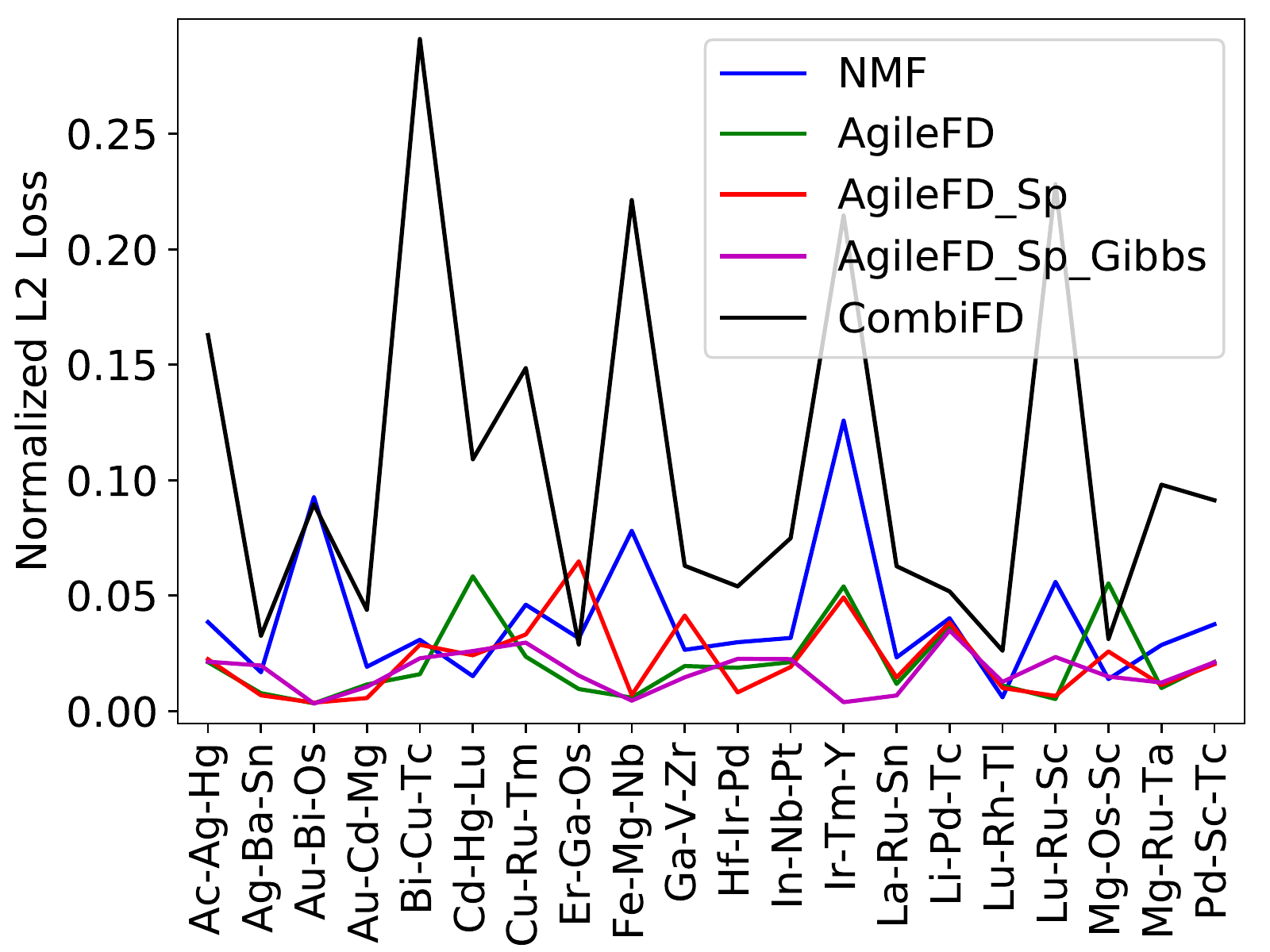}
  \caption{Normalized L2 Loss vs. ground truth phases for NMF, AgileFD, AgileFD with sparsity, AgileFD with sparsity \& Gibbs phase rule and CombiFD for 20 physical systems.  AgileFD, AgileFD with sparsity, AgileFD with sparsity \& Gibbs phase rule perform best.}
  \label{fig:L2matching}
\end{figure}
\begin{figure}[tb]
  \centering
  \includegraphics[width=0.7\linewidth]{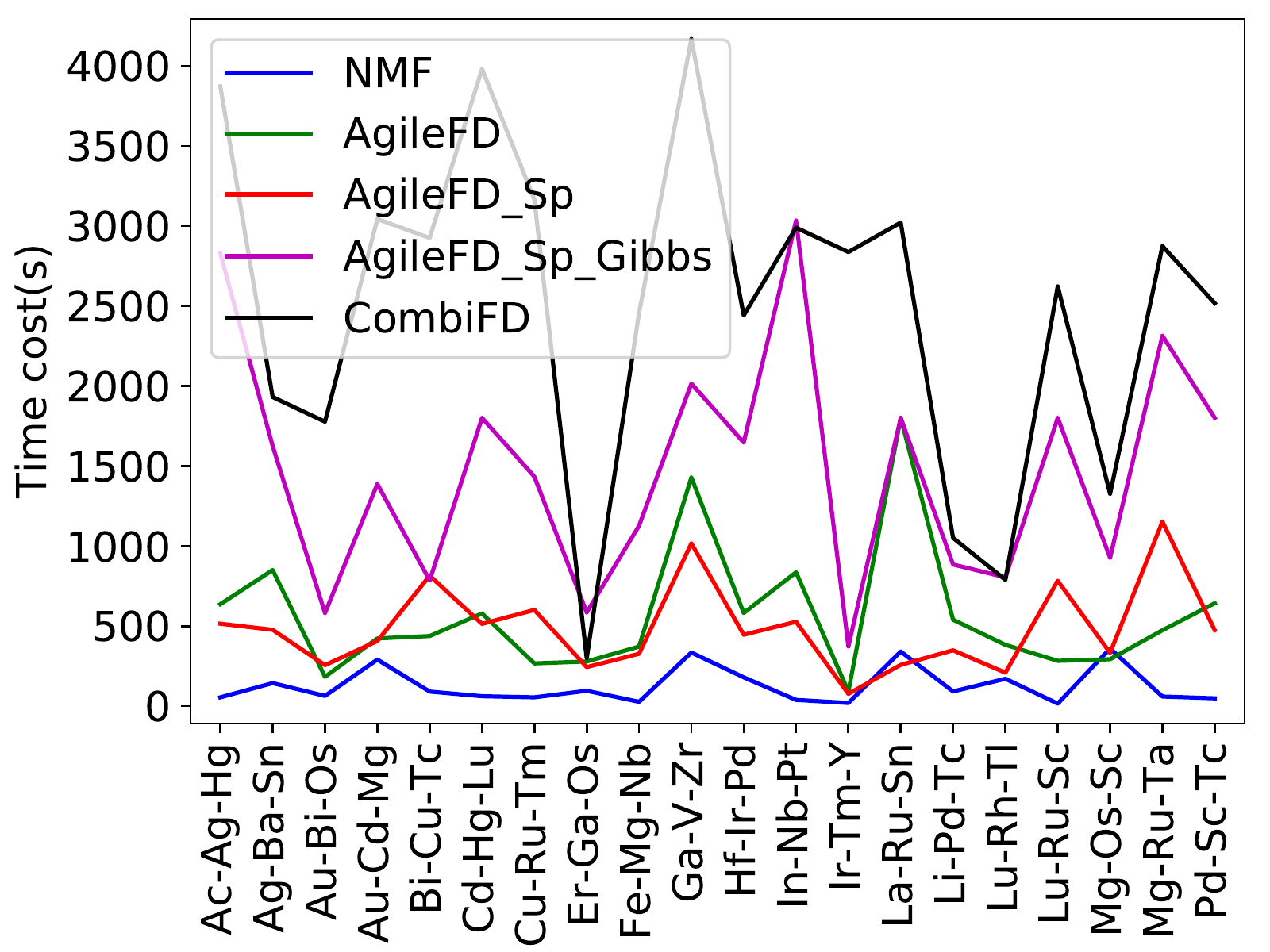}
  \caption{Runtime for NMF, AgileFD, AgileFD with sparsity, AgileFD with sparsity \& Gibbs phase rule and CombiFD to solve 20 physical systems. Times for CombiFD are for up to 15 iterations. The times for the other solvers are until convergence (convergence gap $2\times 10^{-5}$). A time limit of 1 hour was imposed for all solvers. }
  \label{fig:Time}
\end{figure}

\begin{table}[t]
\footnotesize
\renewcommand\tabcolsep{2.75pt} 
\begin{tabular*}{\linewidth}{c|c|c|c|c|c|c}
\hline
System & K & NMF & AgileFD & AgileFD & AgileFD & CombiFD\\
 & & & & Sp &  Sp Gibbs & \\
\hline
Ac-Ag-Hg & 5 & 0.35 & 0.28 & 0.43 & \textbf{1.00} & 0.00 \\
Ag-Ba-Sn & 13 & 0.09 & 0.09 & 0.21 & \textbf{1.00} & 0.00 \\
Au-Bi-Os & 4 & 0.81 & 0.77 & 0.81 & \textbf{1.00} & 0.57 \\
Au-Cd-Mg & 12 & 0.12 & 0.12 & 0.20 & \textbf{1.00} & 0.38 \\
Bi-Cu-Tc & 3 & 1.00 & 1.00 & 1.00 & \textbf{1.00} & 1.00 \\
Cd-Hg-Lu & 7 & 0.12 & 0.11 & 0.22 & \textbf{1.00} & 0.65 \\
Cu-Ru-Tm & 6 & 0.32 & 0.27 & 0.62 & \textbf{1.00} & 0.59 \\
Er-Ga-Os & 8 & 0.08 & 0.12 & 0.43 & \textbf{1.00} & 0.31 \\
Fe-Mg-Nb & 4 & 0.73 & 0.78 & 0.89 & \textbf{1.00} & 0.85 \\
Ga-V-Zr & 13 & 0.10 & 0.07 & 0.41 & \textbf{1.00} & 0.51 \\
Hf-Ir-Pd & 8 & 0.27 & 0.38 & 0.51 & \textbf{1.00} & 0.43 \\
In-Nb-Pt & 10 & 0.14 & 0.14 & 0.34 & \textbf{1.00} & 0.00 \\
Ir-Tm-Y & 4 & 0.86 & 0.94 & 0.98 & \textbf{1.00} & 0.54 \\
La-Ru-Sn & 12 & 0.12 & 0.09 & 0.32 & \textbf{1.00} & 0.63 \\
Li-Pd-Tc & 8 & 0.25 & 0.48 & 0.49 & \textbf{1.00} & 0.53 \\
Lu-Rh-Tl & 9 & 0.17 & 0.12 & 0.36 & \textbf{1.00} & 0.00 \\
Lu-Ru-Sc & 4 & 0.51 & 0.95 & 0.95 & \textbf{1.00} & 0.00 \\
Mg-Os-Sc & 7 & 0.38 & 0.35 & 0.57 & \textbf{1.00} & 0.45 \\
Mg-Ru-Ta & 8 & 0.22 & 0.20 & 0.27 & \textbf{1.00} & 0.00 \\
Pd-Sc-Tc & 9 & 0.24 & 0.14 & 0.44 & \textbf{1.00} & 0.38 \\
\hline
\end{tabular*}
\vskip -8pt
\caption{Percentage of sample points that satisfy the Gibbs Phase Rule
  for each solver on 20 physical systems. Phases that account for less than
 1\% of the modeled signal are not counted towards this phase limit.  }
\label{tab:gibbs}
\end{table}

Despite several solvers having been proposed \cite{LeBras11,long07,kusne14,ermon2012smt,combiFD15}  in recent years, most solvers
were configured and tested only on a handful of systems.
As one contribution of this paper, we provide a large-scale evaluation
of various solvers on the phase-mapping problem.

We generated synthetic ternary metallic systems using data provided by the Materials Project~\cite{Jain2013}, which provides theoretical crystal structure information using Density-Functional Theory based convex hull construction in composition, energy space that predicts compositions and corresponding lowest energy of formation atomic configurations that form the vertices of the convex hull.

The XRD patterns at compositions between the vertices of the convex hull are calculated, using pymatgen~\cite{Ong2013}, by interpolation of a) XRD patterns of the vertex phases or b) modified XRD patterns of the vertex phases to accommodate alloying. We applied a stylized model of solid solubility and alloying, and used structure interpolation to simulate modified phase diagrams that include the additional degrees of freedom from alloying. We calculated XRD patterns for each modified constituent, including their interpolated structures, and combined them according to the mixture proportions in the phase diagram.


We selected 20 examples containing varying numbers of phases and amounts of alloying for our experminent. These data reflect many properties of real experimental data including physically plausible combinations of basis XRD patterns, connectivity relationships of phase fields, adherence to Gibbs phase rule, and peak shifting. These data also provide ground truth information, so that we can directly compare our solution models to those that generated the data.

We compare the resulting phases found by different solvers against the
ground truth phases, which is known beforehand from the simulation used
to generate these 20 systems.  We tested 
NMF\footnote{For consistency, we implemented NMF using generalized KL-divergence loss, equivalent to AgileFD with $M=1$. Because of this, and the resulting use of log scaling of $q$, this implementation of NMF differs from that previously applied to the phase mapping problem~\cite{long09}.}, AgileFD, AgileFD with Sparsity Regularization,
AgileFD with Sparsity and Gibbs phase rule enforced, and CombiFD.  The
convergence gap for NMF and AgileFD are set to be $2\times 10^{-5}$, the MIP gap of CombiFD is set to be 0.2, and
the sparsity parameter is set to 0.35 in AgileFD. All solvers were given a time limit of 1 hour, and allocated one compute core.

First we evaluate the solution quality by comparing the modeled signal for each phase at 
each sample point, including shift, to the known signal from that phase at that sample point. 
We find the permutation of the phases in the solution to best match the ground truth, 
and calculate the L2 loss for each component. These are summed over all phases and samples, 
and scaled by the total value of all signals in the example system.

%

As shown in Figure~\ref{fig:L2matching}, in general the solutions found
by AgileFD (including AgileFD-sparsity and AgileFD-sparsity-gibbs)
better match the ground truth when compared with NMF and CombiFD.
%
%
NMF underperforms because it cannot model peak shifting. CombiFD often only completed a few
iterations within the time limit, and in a few cases timed out in the first iteration. These timeouts before
convergence resulted in lower solution quality. Figure~\ref{fig:Time} compares the runtimes on each
instance.

AgileFD with sparsity regularization and/or Gibbs phase rule enforced 
outperforms AgileFD without extensions. They are able to find solutions that
better match the physical constraints.  We calculate the percentage of
sample points that satisfy the Gibbs phase rule, for the solutions
found by each solver for each instance.
The result is shown in Table~\ref{tab:gibbs}.When
sparsity regularization (or, by design, the Gibbs phase rule) is enforced, the
solutions tend to better satisfy the Gibbs phase rule.





\subsection{Case Study: Nb-V-Mn Oxides}

\begin{figure*}[tp]
  \centering
  \includegraphics[width=0.9\linewidth]{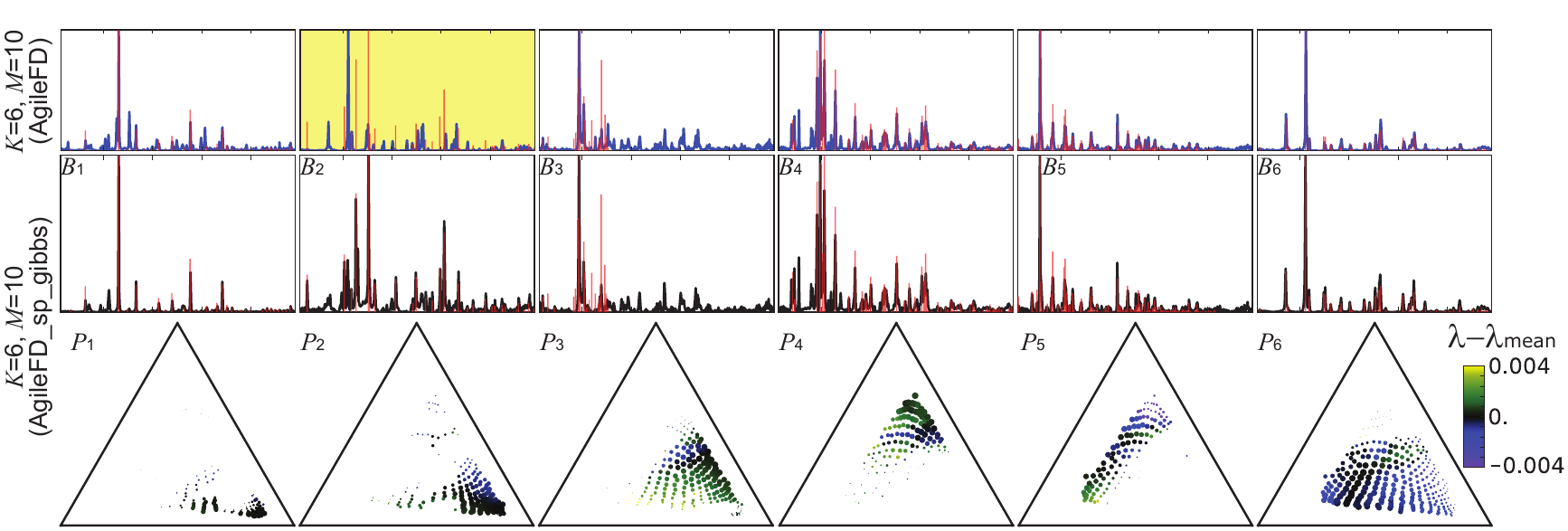}
  \vskip -10pt
  \caption{The $K$=6 basis patterns for the AgileFD solution (top) and AgileFD-sparsity-gibbs solution (middle) for the Nb-V-Mn system are shown along with stick patterns (translucent red) of the crystal structures identified using the AgileFD-sparsity-gibbs solution. The primary discrepancy in the AgileFD solution is highlighted in yellow. The phase map representation of $H$ is shown as a series of composition plots for the AgileFD-sparsity-gibbs solution (bottom). For each basis phase, the composition plot shows the samples containing the phase with point size indicating the concentration of the phase and the point color indicating the fractional shift with respect to the plotted basis patterns. The elemental labels for the composition diagrams are shown in Figure \ref{fig:numphases}.}
  \label{fig:NbVMn_soltn}
\end{figure*}
\vskip -5pt

The integration of the rapid solver with visualization tools enables materials scientists to take advantage of the tunable initialization and constraint parameters to create a meaningful solution. An illustrative example is found in a ternary composition library containing a broad range of compositions in the Nb-V-Mn oxide space. While the phase behavior of binary sub-compositions (e.g. Nb-V oxides) have been previously studied, the ternary compositions are being explored for the first time to discover solar light absorbers for energy applications. While the set of 317 XRD patterns provides sufficient information to solve the phase behavior of these oxides, the materials researchers were unable to obtain a meaningful phase diagram for over a year due to the complex phase behavior that evaded comprehension via manual analysis. Using the visualization tools and rapidly trying various solutions, the researchers determined that there are $K$=6 primary phases in this dataset. For this case study, we compare solutions from 3 different solvers with $K$=6: (1) NMF; (2) AgileFD with $M$=10, which includes shifting of each basis pattern by up to approximately 2\%; (3) AgileFD-sparsity-gibbs with $M$=10. 

\begin{figure}[tp]
  \centering
  \includegraphics[width=0.95\linewidth]{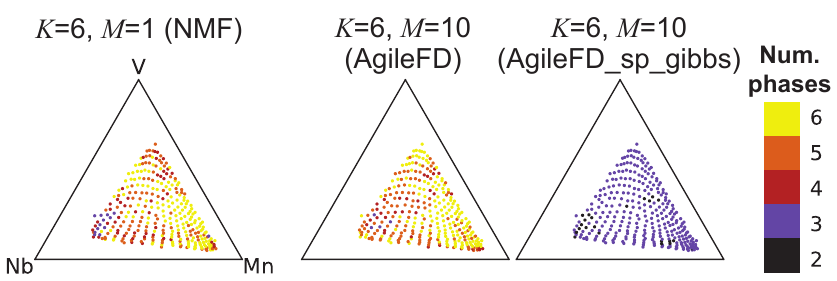}
  \caption{Composition maps of the number of the $K$=6 basis components utilized in the solutions from 3 different solvers: NMF, AgileFD, and AgileFD-sparsity-gibbs. Values in excess of 3 are non-physical.}
  \label{fig:numphases}
\end{figure}

As described above, adherence to the Gibbs phase rule is important for providing the researcher with a meaningful phase diagram. The number of basis patterns utilized for each sample is shown in Figure \ref{fig:numphases} for each of the 3 solutions. The NMF solution adheres to the Gibbs phase rule for only ~4\% of the samples, with nearly half of the samples utilizing all 6 basis patterns. While the AgileFD solution is more meaningful than the NMF solution due to the tracking of alloying via basis pattern shifting, the AgileFD solution uses more phases for several compositions, corresponding to a worse violation of Gibbs phase rule. This property of the AgileFD solver may be understood intuitively by considering that for a given sample, any of the $M$ copies of each basis pattern can be utilized to model small features in the XRD patterns, resulting in small amounts of shifted patterns to appear across the composition space and further motivating the encoding of Gibbs phase rule in the solver. With this constraint, the AgileFD solution utilizes a maximum of 3 phases and in some composition regions only 2 phases are utilized. 

Figure \label{fig:NbVMn_soltn} summarizes the AgileFD-sparsity-gibbs solution, and the desirable attributes that were not enforced but provide meaning to the researcher include: (1) excellent matching of the primary peaks of each basis pattern with a known crystal structure, demonstrating that each basis pattern is truly representative of a phase, (2) excellent composition space connectivity of each phase concentration map, as expected for equilibrium phase behavior, (3) systematic compositional variation in the shift parameter $\lambda$, demonstrating alloying within the phases, in particular phases 3, 5, and 6. While the AgileFD solution includes pattern shifting, the lack of adherence to the Gibbs phase rule has important consequences on the solution that cannot be ameliorated through modification of the $H$ to impose Gibbs phase rule \textit{ex post facto}, without a corresponding refinement of $W$. The most prominent difference is highlighted in Figure \ref{fig:NbVMn_soltn} where the second basis pattern of the AgileFD basis pattern is quite different from that of the AgileFD-sparsity-gibbs solution. This AgileFD basis pattern contains a mixture of signals from other phases and is thus not able to be matched to a known structure. The researcher would thus be left with the impression that this may be a newly discovered phase and undergo the arduous task of hypothesizing new crystal structures that could correspond to this basis pattern. The false-positive phase discovery from AgileFD is circumvented in the AgileFD-sparsity-gibbs solution as the encoding of Gibbs phase rule in the solver results in effective de-mixing of basis patterns such that all 6 primary phases are identified. Indeed, only by imposing \textit{a priori} constraints in AgileFD is a complete, meaningful solution produced.

\section{Conclusion}

High-throughput materials discovery is revolutionizing the efficiency of materials science experiments through automated characterization of thousands of materials in a single library.  A major, critically-missing component of the high throughput materials discovery pipeline is the ability to rapidly solve the phase map identification problem, which involves the determination of the underlying phase diagram of a family of materials from their composition and structural characterization data. To address this substantial challenge, we developed Phase-Mapper, a comprehensive AI platform that tightly integrates XRD experimentation, AI problem solving, and human intelligence for solving the phase mapping problem. AI solvers in Phase-Mapper provide, within minutes, high-quality solutions to the phase mapping problem, which can then be examined and further refined by materials scientists, interactively and in real time.  The de-mixing algorithm is a cornerstone of the Phase-Mapper platform, and previously-developed algorithms routinely produce non-physical solutions. We have developed a novel solver, AgileFD, that features lightweight iterative updates of candidate solutions and a suite of novel adaptations to the multiplicative updating rules. In particular, we have developed the ability to incorporate constraints that capture the physics of materials as well as human feedback, enabling functionalities well beyond traditional de-mixing techniques and producing physically-meaningful solutions.  We compare different solver variants with previously proposed methods in a large-scale experiment involving 20 synthetic systems, demonstrating the efficacy of imposing physical constrains using AgileFD, while keeping fast solution times. Phase-Mapper has been deployed in the Joint Center for Artificial Photosynthesis (JCAP) supported by the Department of Energy for materials scientists to solve a wide variety of real-world phase diagrams, including the previously unsolved Nb-Mn-V oxide system, which is provided here as a case study and an illustrative example of the benefits of encoding physical constraints. We believe Phase-Mapper will lead to further developments in high-throughput materials discovery by providing rapid and critical insights into the underlying phase behavior of new materials.

\section*{Acknowledgements}
This material is supported by NSF awards CCF-1522054 and CNS-0832782 (Expeditions), CNS-1059284 (Infrastructure), and IIS-1344201 (INSPIRE); and ARO award W911-NF-14-1-0498. Materials experiments are supported through the Office of Science of the U.S. Department of Energy under Award No. DE-SC0004993. Use of the Stanford Synchrotron Radiation Lightsource, SLAC National Accelerator Laboratory, is supported by the U.S. Department of Energy, Office of Science, Office of Basic Energy Sciences under Contract No. DE-AC02-76SF00515.

{\small
\bibliographystyle{aaai}
\bibliography{bibs/MDbib,bibs/MDbib2,bibs/NMF,bibs/john,bibs/rich}  
}

\end{document}